# The MBPEP: a deep ensemble pruning algorithm providing high quality uncertainty prediction

Ruihan Hu, Qijun Huang[*], Sheng Chang, Hao Wang and Jin He

**Abstract** Machine learning algorithms have been effectively applied into various real world tasks. However, it is difficult to provide high-quality machine learning solutions to accommodate an unknown distribution of input datasets; this difficulty is called the uncertainty prediction problems. In this paper, a margin-based Pareto deep ensemble pruning (MBPEP) model is proposed. It achieves the high-quality uncertainty estimation with a small value of the prediction interval width (MPIW) and a high confidence of prediction interval coverage probability (PICP) by using deep ensemble networks. In addition to these networks, unique loss functions are proposed, and these functions make the sub-learners available for standard gradient descent learning. Furthermore, the margin criterion fine-tuning-based Pareto pruning method is introduced to optimize the ensembles. Several experiments including predicting uncertainties of classification and regression are conducted to analyze the performance of MBPEP. The experimental results show that MBPEP achieves a small interval width and a low learning error with an optimal number of ensembles. For the real-world problems, MBPEP performs well on input datasets with unknown distributions datasets incomings and improves learning performance on a multi task problem when compared to that of each single model.

## 1. Introduction

Machine learning has been successful in many areas such as memristive designing [1], conversational system [2], ECG signal recognition [3], autonomous vehicles [4] and brain-inspired network construction [5-7]. Normally, the data distributions in testing datasets should be get close to those of the training datasets, if good learning results to be obtained. However, in many situations that are instances of the uncertainty prediction problem, such as in obtaining a skin cancer diagnosis [8], dog breed classification [8], lesion segmentation [9], reinforcement learning [10], noisy data uncertainty [11] and model uncertainty [8] [12]; the distributions of test datasets are unknown. In these cases, it is difficult for traditional learning models to achieve good performance.

Qijun Huang
Wuhan University, Wuchang District, Wuhan 430072, Hubei, China
Tel.: +86-13387542819
E-mail: huangqj@whu.edu.cn

Some researchers used Bayesian neural networks that depended on the prior distribution of the datasets [10] [13], and Markov chain Monte Carlo [10] and [14] to solve the uncertainty prediction problems. However, such approaches are computationally expensive when they are applied into the large scale networks. To avoid the prior distributional assumption, we deal with uncertainty prediction as a prediction intervals (PI) problem [15], which uses upper and lower bounds to quantify the degree of uncertainty. The previous studies [15] [16] [17] redefined the prediction intervals into the two targets, namely, the minimum being the width between the upper and lower bounds, and the maximum being the prediction interval coverage probability. As the loss functions in the above studies are nonconvex according to [16], the prediction intervals cannot be calculated by the standard gradient descent method, which increases the complexity of solving this kind of problems. To overcome this difficulty, we design a simple and intuitive loss function that is unified and convex behind the last layer of the network; the network can be learned with the standard gradient descent.

Furthermore, due to the high degree of uncertainty degree of outputs, it is difficult to assess the PI performance by using only one neural network, so combining several neural networks into an ensemble one becomes a suitable method of solving the PI problem. Unfortunately, to the best of our knowledge, no previous study has linked the optimal number of ensembles with the PI. In this manuscript, we introduce a Pareto pruning method based on margin criterion fine-tuning (MCFT) to tackle this problem.

Overall, the margin-based Pareto deep ensemble pruning (MBPEP) model is built on the deep models; and these models are ensemble-pruned to account for the prediction's uncertainty in this manuscript. The remainder of this paper is organized as follows. Section **2.1** briefly discusses the unique loss functions in each sub-learner. Section **2.2** introduces the Pareto pruning method based on margin criterion fine-tuning. Section **2.3** describes the entire learning procedure of MBPEP. In Section **3**, a series of experiments are used to quantify the loss functions and the ensemble pruning. MBPEP is tested on some real-world benchmarks in Section **4**. Conclusions are given in Section **5**.

## 2. Materials and Methods

### 2.1 Unique loss function for prediction intervals

We treat the PI problems as prediction intervals [15] processes that do not depend on any prior assumption. The learned intervals are narrow, and the output's uncertainty is accounted for. Let the inputs $X$ be targeted by $Y$; and the upper and lower bounds of the prediction are denoted by $\hat{Y}_L$ and $\hat{Y}_U$ in this manuscript. The prediction interval coverage probability ( $PICP$ ) and mean prediction interval width ( $MPIW$ ) are the essential metrics in $PI$ ; and their usual mathematical formulas are as follows:

$$PICP = \frac{1}{N}\sum_{i=1}^{N} k_i \tag{1}$$

$$MPIW = \frac{1}{N}\sum_{i=1}^{N}(\hat{Y}_U(X_i) - \hat{Y}_L(X_i)) \tag{2}$$

where $N$ denotes the number of samples, and $k_i$ denotes the indicator ($k_i = 1$ if $\hat{Y}_L(X_i) \leq Y_i \leq \hat{Y}_U(X_i)$; otherwise, $k_i = 0$). According to a previous work [18], the values of *PCIP* will be decreased if *MPIW* narrows. Unfortunately, these two metrics are usually difficult to be keep in balance. In this manuscript, this restriction is slightly alleviated by allowing some of targets $Y$ to not fall into the range between $\hat{Y}_U$ and $\hat{Y}_L$ to some degree in Eq. (2). Thus, the *MPIW* can be represented as follows:

$$MPIW = \frac{1}{N}\sum_{i=1}^{N}(\hat{Y}_U(X_i) - \hat{Y}_L(X_i))k_i \tag{3}$$

In what follows, to allow *PICP* to remain at a reasonable value, the confidence level $1-\phi$ is defined (where $\phi$, namely denotes the deviation level) as the degree of tolerance; the corresponding output *PICP* should be greater than $1-\phi$. According to the description of LUBE [15], *MPIW* is contained in the loss function to obtain a smaller value. Additionally, $-PICP$ is obtained by maximizing the corresponding regularization term as follows:

$$Loss_{LUBE} = \frac{MPIW}{\hat{Y}_U - \hat{Y}_L}(1 + \exp(c\max(0,(1-\phi) - PICP))) \tag{4}$$

where $c$ denotes the penalty coefficient, and the regularization term has the exponential form. The loss function is designed to minimize the *MPIW* while keeping the value of *PICP* greater than confidence level $1-\phi$ in the meanwhile.

Normally, similarly to the *PI* framework of LUBE in Eq.(4), the loss function is nonconvex or nondifferentiable. The weights and bias of the neural networks may be trapped in local minima when gradient descent optimization and an evolutionary algorithm are used, such as employing the simulated annealing to solve Eq. (4). To design a computation framework that can avoid the nonconvex problem of computing $\hat{Y}_U$ and $\hat{Y}_L$, more nonlinear activation operators are added in the MBPEP. These operators can ensure the differentiability and convexity in each sub-learner; thus, those sub-learners can be easily solved by the gradient descent methods such as stochastic gradient descent, AdaDelta [19], Adam [20], etc. The complete feed-forward architecture for the base learner of MBPEP is shown in Fig. 1.

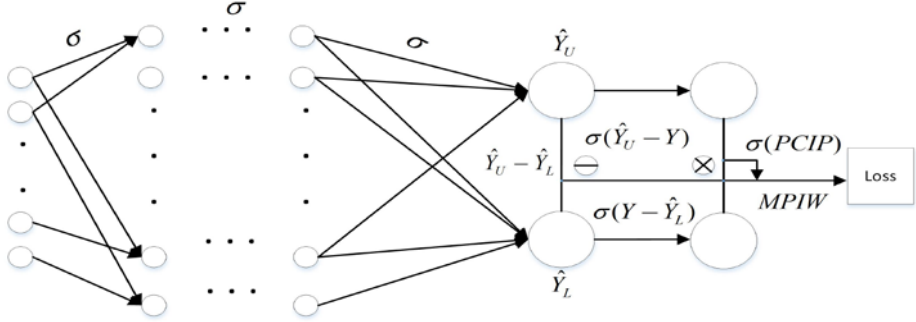

**Fig. 1**: Feed-forward network structure of the base leaner of MBPEP. It consists of the deep layer architecture and outputs the loss determinated by the $MPIW$ and $PICP$.

In Fig. 1, $\sigma$ denotes the standard activation functions (such as the Sigmoid and Relu, which determined by the architecture of the network). When the layer architecture of the network is not very deep, the Sigmoid function is used as the activation operator. As to the deep layer architecture, the Relu function is used due to its good gradient propagation characteristics [21]. To avoid over-fitting phenomenon, a dropout mechanism [22] is added in each layer by randomly selecting units to be dropped. The fixed retention probability of dropout is set to 0.8. Before the upper and lower bounds $\hat{Y}_U$ and $\hat{Y}_L$, the structure of the base leaner can be replaced by the standard forms such as feed-forward, recurrent and convolutional layers, according to various real-world problems. To allow the weights in the base learner to converge, the indicator $k_i$ in Eq.(1) is effectively qualified, and the activation functions are added on the difference between the upper bounds, lower bounds and targets $Y$, as follows:

$$k_i = \sigma(\hat{Y}_U - Y) \otimes \sigma(Y - \hat{Y}_L) \tag{5}$$

where $\otimes$ denotes the elementwise product operator. Consistently with (1), our method's mean prediction interval width $PICP_{MBPEP}$ can be redenoted as $\frac{1}{N}\sum_{i=1}^{N} k_i$. Similarly, to the calculation of $MPIW$ in Eq.(3), the differences of upper and lower bounds $\hat{Y}_U(X_i) - \hat{Y}_L(X_i)$ are kept the same in MBPEP, as they are computed by the non-linear operators in the last layer of the neural network. The new mathematical expression for our $MPIW_{MBPEP}$ is:

$$MPIW_{MBPEP} = \frac{1}{N}\sum_{i=1}^{N}(\hat{Y}_U(X_i) - \hat{Y}_L(X_i))(\sigma(\hat{Y}_U(X_i) - Y_i) \otimes \sigma((Y - \hat{Y}_L))) \tag{6}$$

From Eq. (6), it can be seen that the new $MPIW$ can be simplified and kept the same as in the original framework: $MPIW_{MBPEP} = \frac{1}{N}\sum_{i=1}^{N}(\hat{Y}_U(X_i) - \hat{Y}_L(X_i))k_i$. It is worth noting that to allow the derivation of the loss function derivable, the hinge regularization term is introduced in the loss function (Eq.(4)) of our MBPEP method that can be easily converted to the product of the penalty constant $c$ and a nonlinear $Relu$ operator, the derivative of which can be easily computed:

$$Loss_{MBPEP} = MPIW_{MBPEP} + c \cdot Relu((1-\phi) - PICP_{MBPEP}) \tag{7}$$

## 2.2 Ensemble pruning

In Section 2.1, we have introduced the architecture of the base learner in MBPEP, but there is still a problem of determining the number of base learners are needed for achieving good performance. In the previous studies ([15], [18], [23-24]), the numbers of the base learners were pre-defined, and researchers did not relate the learning performance to the number of the ensembles. In this section, we further explore the particular metric of measuring the ensemble performance and determining the relationship between the learning performance and the number of ensembles.

In this manuscript, the Pareto evolution [25] method as a bi-objective evolutionary algorithm is used to deal with this problem. The training losses of base learners described in Eq. (7) are used to measure the learning performance. Moreover, considering that the differences of the sub-learners' outputs may be vary, the learners that output narrow boundary intervals are selected first in the ensembles. To this end, margin criterion fine-tuning (MCFT) is added to measure the performance of each sub learner. Assuming that there are $T^*$ base learners $H^* = \{H_1.....H_{T^*}\}$ in the ensembles, $\hat{Y}_U^t(x_i)$ denotes the upper bound with respect to $i_{th}$ inputs $x_i$ that is determined by $t_{th}$ sub-learner, and $\hat{Y}_L^t(x_i)$ means the corresponding lower bound. The margin for $x_i$ is defined as the mean value between the upper and lower bounds among the ensembles $H^*$, and can be represented as follows:

$$m\arg in(x_i) = \frac{1}{T^*}\sum_{t=1}^{T^*}\left|\hat{Y}_U^t(x_i) - \hat{Y}_L^t(x_i)\right| \qquad (8)$$

Next, we score the margin of the ensembles of all datasets by:

$$C_{H^*}(X) = \frac{1}{N}\sum_{i=1}^{N}\log[m\arg in(x_i)] \qquad (9)$$

where $N$ denotes the number of the training samples. According to Eq. (8), it is observed that if the output boundary intervals of those base learners are quite large, the corresponding margin criterion score $C_{H^*}(X)$ will be high. The learning performance of the ensembles can be described as the summation of the losses (according to Eq. (7)) and their margin criterion score $C_{H^*}(X)$:

$$f(H^*) = C_{H^*}(X) + Loss \qquad (10)$$

As the learning performance and the number of sub-learners are two objects in the optimization space, we treat this problem as a bi-objective programming formulation and minimize it as follows:

$$\arg\min_{s\in\{0,1\}^T} (f(H^*), sum(s)) \qquad (11)$$

where $s$ denotes the chosen indicator ($s_t = 1$ if a sub-learner $H_t$ is chosen; otherwise, $s_t = 0$). Thus $sum(s)$ represents the number of the ensemble size. Unlike other single-objective algorithms [26-27], the 'domination' conception is introduced, which disturbs the ensemble size continuously and measures the difference between $f(H^*)$ and $sum(s)$ during the iteration process.

## 2.3 Margin-based Pareto deep ensemble pruning (MBPEP) algorithm

In this section, the complete learning pipelines of MBPEP are described as follows:

Step 1: Data sampling.

At the beginning of the training process, the dataset is randomly re-sampled by the classical bootstrap. The number of the bootstrap repetitions is predefined at the start of training.

Step 2: Parameter initialization.

At this step, the configurations of MBPEP is initialized, such as the hidden size, the type of the activation functions, the number of the batches, the maximum number of the epochs, and the maximum number of the base learners, is initialized. In the following Sections 3 and 4, in order to measure the performances of MBPEP and the other state-of-the-art models, some of these hyper parameters are kept fixed. In particular, the number of outputs for each sub learner should be fixed at two, which represents the upper and lower bounds.

Step 3: Training.

Assume that $H_T$ and $H_{T^*}$ represent the original and optimal ensemble pools, where $H_{T^*} \subseteq H_T$. The architectures of the base learners have been discussed in deail in Section 2.1; and each of them is trained independently with the gradient descent method to obtain the optimal weights and biases. Due to the scalable computation framework in Fig. 2, in order to measure the uncertainty prediction performance of different loss functions in Section 3.2.1, the loss functions in this step need to be changed, and the other steps need to remain fixed.

Step 4: Ensemble pruning.

A number of the base learners are quite redundant. To save storage space and accelerate prediction speed, they need to be pruned. The ensemble process has been discussed in detail in Section 2.2; the optimal base leaners are selected to construct the new ensembles. To measure the pruning performance of MBPEP, the pruning process should to be combined with other algorithms while the other steps remain fixed.

Step 5: Integration.

In the integration phase, we use the simple median voting method to fuse several outputs of optimal ensembles to the final upper and lower bounds. Of course, other nontrainable [28-29] and trainable integration methods [30-31] can also be used.

Step 6: Testing.

The optimized base learners are used to test the uncertainty prediction dataset.

The complete learning process is shown in Fig. 2.

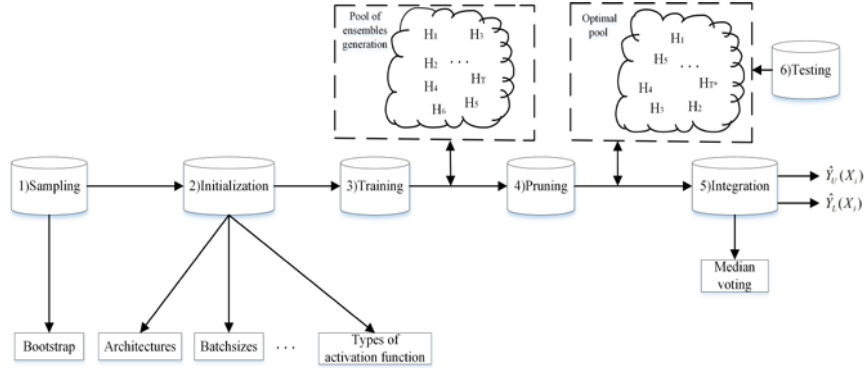

**Fig. 2**: Complete learning process of MBPEP. It contains the sampling, initialization, training, pruning, integration and testing steps. In the training stage, all sub-learners are trained independently, and some of these learners are pruned in the pruning step. Finally, these optimal ensembles are fused by median voting.

## 3. Performance analysis

### 3.1 Configuration

Several UCI datasets [32] are used as benchmarks in this section. The description of these datasets is shown in Tab. 1; each of them is split into three parts; the training, validation and testing datasets. The feed-forward architecture is used for the sub-learners. In Section 3.2, various loss functions and ensemble pruning methods are compared. In Section 3.3, MBPEP is tested to compare its effectiveness to that of several ensemble pruning methods.

**TABLE 1:** UCI DATASETS

| Data description | | | |
|---|---|---|---|
| Dataset | Training size | Validation size | Testing size |
| Adult | 24421 | 9768 | 14653 |
| Arcene | 450 | 180 | 270 |
| Australian | 3325 | 1330 | 1995 |
| Breast-cancer | 285 | 114 | 170 |
| Disorders | 173 | 69 | 103 |
| Heart | 135 | 54 | 81 |
| Wine | 89 | 36 | 53 |
| Ionosphere | 176 | 70 | 105 |
| Kr-vs-kp | 1598 | 639 | 959 |
| Letter | 10000 | 4000 | 6000 |
| Optdigits | 1912 | 760 | 1151 |
| Satimage | 1155 | 462 | 693 |
| Sonar | 695 | 278 | 416 |
| Spambase | 2301 | 920 | 1380 |
| Tic-tac-toe | 479 | 192 | 287 |
| Vehicle | 473 | 189 | 284 |
| Vote | 218 | 87 | 130 |

## 3.2 Comparison experiments

3.2.1 Comparison of various loss functions for uncertainty prediction

In this Section, MBPEP is evaluated to capture the synthetic data that combines of the real data sets with various noise distributions. The example dataset is a one-dimensional toy regression [33], in which the regression formulation can be written as: $y = x^3 + \varepsilon$, where the noise data $\varepsilon$ follows the Gaussian distribution. In this section, we will show the advantage of MBPEP over other uncertainty estimation models, such as the Lower Upper Bound Estimation model (LUBE) [15] and the Concrete Dropout model (CD) [34]. As described in Section 2.1, the LUBE uncertainty estimation model is similar to our model and outputs two boundaries and a loss function constructed by $MPIW$ and $PICP$. However, the loss function of LUBE is nonconvex and can be optimized by the simulated annealing method described in this paper. The CD model outputs two nodes that estimate the mean and variance values of a Gaussian distribution; and negative log likelihood (NLL) is used as the loss function. In contrast, for the MBPEP model, a more unified computation framework that combines the corresponding loss function $L_{MBPEP}$ (composed of $MPIW$ and the hinge loss form of the $PICP$ architecture to predict uncertainty) and Pareto ensemble pruning is used. Additionally, non-pruning MBPEP without ensemble pruning (NP-MBPEP) is used as a baseline to measure performance of a model without the ensemble pruning technique. The results are shown in Fig. 3.

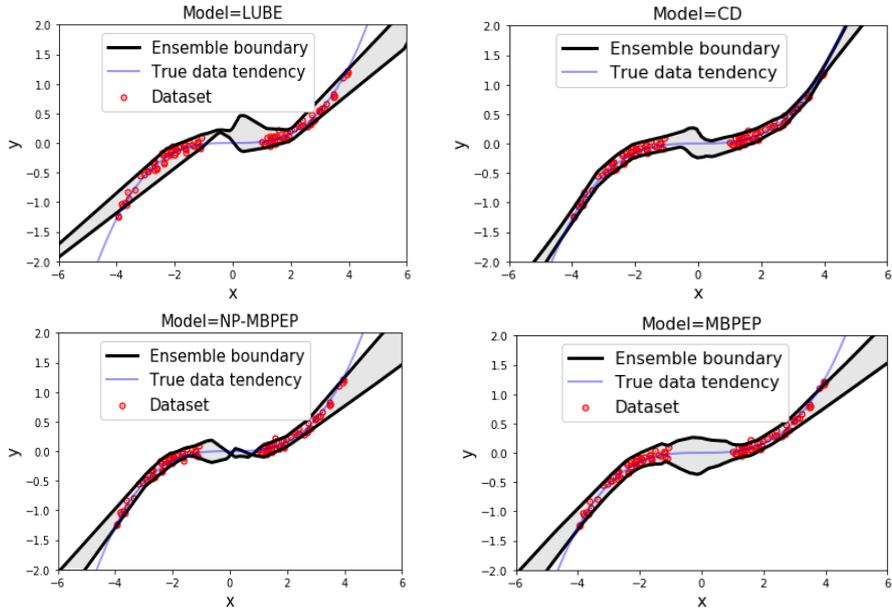

**Fig. 3**: Outputs of (a) LUBE, (b) CD, (c) NP-MBPEP, and (d) MBPEP. The output boundaries assume the Gaussian distribution.

Specifically, $\varepsilon$ is the noise that follows the Gaussian distribution and the confidence level $(1-\phi)$ is set to $0.95$ in this section. As described in Fig. 3, the red points denote the noisy datasets. The gray area between the black thick lines denotes the ensemble PI boundaries that are determined by the upper and lower boundaries. The blue lines denote the ground truth curves. Fig. 3 shows that the PI boundaries of LUBE and NP-MBPEP hardly capture the real truth curves at the inflection points. However, for the CD and MBPEP models, the PI boundaries can perfectly fit the truth curves and cover all of the samples. In particular, for MBPEP, they can restore the inflection regions that cannot be fitted effectively by the none-pruning MBPEP.

As MBPEP does not need to assume in advance the distribution of data samples, unlike the Gaussian distribution of noise data in the preceding paragraph, we further explore the performance of these models applied to samples with exponential function $y = exp(x) + \varepsilon$, where $\varepsilon$ means the exponential noise. The results are shown in Fig. 4.

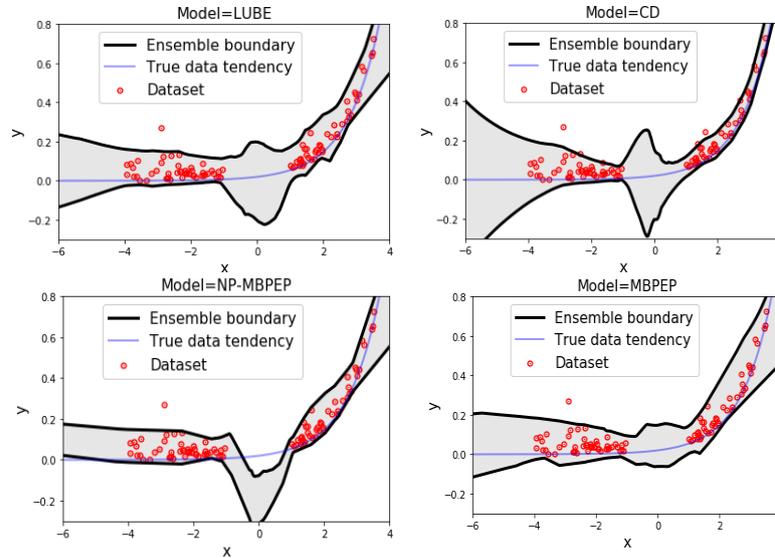

**Fig. 4**: Outputs of (a) LUBE, (b) CD, (c) NP-MBPEP, and (d) MBPEP. The output boundaries are under the exponential distribution.

Fig. 4 shows that the interval width between the upper and lower bounds of model CD is too large, indicating that it cannot effectively capture the other distributions of the data samples other than the Gaussian distribution. As to the other models, the LUBE model can output ideal boundaries except for the point of inflection under the exponential distribution. In contrast, MBPEP can fit the true data tendency most perfectly. The experiments demonstrate that MBPEP is a distribution-free model that can efficiently capture datasets under any distribution.

3.2.2 Comparison of ensemble pruning methods to standard ensembles

In this section, the effect of the ensemble pruning technique will be explored under the optimal number of ensembles, and its performance on uncertainty classification tasks will be examined. To calculate the effectiveness of pruning, the combinations of $[5, 10, 20, 30, 50, 70, 100]$ sub-learners are constructed, and the testing time consuming is used as the metric. Performances of MBPEP and NP-MBPEP is compared in Fig. 5. It is clearly observed that the ensemble pruning technique can reduce the testing time significantly compared to that of NP-MBPEP significanlly, with the increase of the original ensemble size. According to Fig. 5, if there are more sub-learners being trained, a significant amount of time is saved.

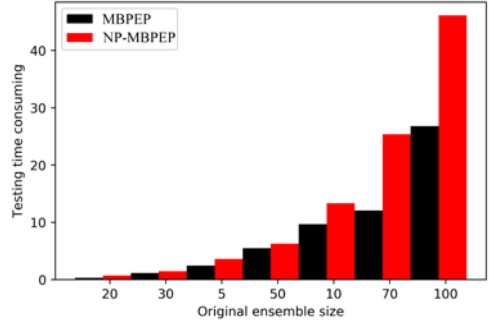

**Fig. 5**: Testing efficiency comparison of MBPEP and NP-MBPEP for various original ensemble sizes. Black and red columns correspond to MBPEP and NP-MBPEP, respectively.

3.2.3 Comparison to other pruning methods

In this section, the characteristics of several ensemble pruning models aiming to choose some competent sub leaners in the pool of the ensembles are explored and compared. For readability, these methods are briefly described as follows:

- Dynamic ensemble selection performance (DES-P) pruning: DES-P [35] tends to select the ensembles that achieve the best competence level on $k$ nearest neighbors on the datasets. In DES-P, the competence of pruning is computed by subtracting the performance of a random classifier from the estimation.
- Kullback-Leibler (KL) pruning: Kullback-Leibler [35], an important metric in information theory, is used to compute the competence. The Gaussian potential function is also used in the Kullback-Leibler pruning.
- Dynamic Frienemy indecision ensemble pruning (DFP) [36] is an ensemble pruning method that focuses on defining the competence region of each dataset. As to the pruning stage, the model detects the indecision region and prunes the classifiers in which the decision boundaries do not cross the level of competence of the test samples. In the dynamic selection stage, DFP chooses the classifiers that were preprocessed by the pruning stage.
- Margin Distance Minimization (MDM) pruning: The margin distance [37] defines the quantity $c$ between outputs of ensemble classifiers and labels. The Quantity $c$ equals to 1 if the label is correctly classified. Otherwise, it is equal to -1 if the label is misclassified. The signature vector of ensemble classifier is computed by the summation of $c$ for the selected classifiers. This method is aimed to explore the optimized ensemble classifiers that ensure that all $c$ have positive values. The distance $p$ of Margin Distance Minimization pruning is set to 0.075 in this manuscript.
- Meta-pruning [37] regards the ensemble pruning problem as a meta-problem, and different meta-features are extracted to train the meta-classifier. Meta-features that are used to estimate the competence level of the base learners demonstrate more robust pruning results. The meta-pruning rule is to learn whether a base learner has enough competence level for the incoming datasets.
- Meta-dynamic ensemble selection. Oracle (META-DES.O) [38] is inspired by the meta-pruning based on the Oracle. In contrast to meta-pruning, meta-feature selection is optimized by the Binary Particle Swarm Optimization (BPSO). Afterwards, the meta-classifiers are integrated by the majority voting scheme.

**TABLE 2**: Test errors of ensemble pruning methods

| Datasets | Test error(%) | | | | | | |
|---|---|---|---|---|---|---|---|
| | **Ours** | DES-P | KL pruning | DFP | MDM pruning | Meta pruning | META-DES.O |
| Adult | 17.22 | 20.04 | 31.56 | 22.02 | 17.67 | 20.16 | 19.38 |
| Arcene | 15.73 | 33.33 | 20.83 | 33.33 | 39.10 | 21.21 | 30.30 |
| Australian | 15.68 | 17.98 | 14.76 | 18.42 | 14.80 | 15.53 | 13.15 |
| Breast-cancer | 27.43 | 24.21 | 29.80 | 26.31 | 28.70 | 28.82 | 21.05 |
| Disorders | 30.40 | 32.45 | 36.58 | 35.08 | 33.71 | 31.16 | 31.50 |
| Heart | 20.43 | 19.58 | 23.54 | 24.74 | 21.40 | 19.93 | 23.71 |
| Wine | 1.13 | 1.69 | 1.69 | 3.38 | 1.69 | 1.69 | 4.3 |
| Ionosphere | 4.35 | 7.75 | 4.31 | 7.75 | 4.31 | 6.03 | 4.32 |
| Kr-vs-kp | 5.32 | 5.02 | 6.16 | 5.87 | 6.16 | 5.68 | 6.63 |
| Letter | 16.65 | 20.75 | 17.24 | 21.39 | 18.48 | 22.80 | 26.53 |
| Optdigits | 3.57 | 4.67 | 9.16 | 5.30 | 3.58 | 3.72 | 4.83 |
| Satimage | 10.83 | 11.44 | 12.33 | 12.80 | 12.33 | 12.57 | 24.81 |
| Sonar | 24.84 | 25.63 | 31.06 | 31.88 | 24.95 | 25.70 | 28.98 |
| Spambase | 6.54 | 7.70 | 9.33 | 7.70 | 6.60 | 6.68 | 7.57 |
| Tic-tac-toe | 18.13 | 20.88 | 21.26 | 24.05 | 20.88 | 23.41 | 20.88 |
| Vehicle | 22.44 | 21.78 | 25.73 | 30.71 | 23.36 | 24.42 | 20.14 |
| Vote | 4.36 | 4.23 | 4.64 | 4.72 | 4.15 | 4.32 | 4.44 |
| **Count of wins** | 10 | 1 | 1 | 0 | 2 | 0 | 3 |
| **Direct wins** | | 12 | 15 | 16 | 14 | 14 | 13 |

In this section, in order to measure the performance of margin criterion fine-tuning (MCFT) in MBPEP, ensemble pruning methods that have been described in preceding paragraph are benchmarked on 17 datasets from the UCI datasets. Each result is recomputed under 30 times. To intuitively show the performance of each ensemble pruning method, we first measure the testing errors on the uncertainty classification tasks, shown in Tab. 2. According to Tab. 2, MCFT shows the best performance on 10 datasets (58.82%) of the total number of datasets, which is superior to results of other ensemble pruning methods. The reason is that MCFT can effectively search the sub learners, which maintains the low training losses and the margin criterion score (Eq. 10). Tab. 2 shows that our ensemble pruning method can effectively improve classification performance for uncertainty prediction problems.

3.3 Optimal number of the sub learners

To obtain the optimal ensembles to predict uncertainty, optimal ensemble size is an important metric in ensemble pruning methods. The optimized ensembles' performance on the uncertainty classification tasks is shown in Tab. 3. It is easily observed that MCFT achieves the smallest ensemble sizes on the 58.8% (10/17) sets of all data sets, while the other methods do so on less than 41.2% (7/17) of datasets. Tab. 3 also demonstrates that MCFT can be less time-consuming than are other ensemble pruning methods at the testing stage. Tab. 3 shows that our ensemble pruning method can effectively shrink the number of sub-learners with uncertainty predictions.

**TABLE 3**: Optimal ensembles among ensemble pruning methods

| Datasets | Optimal ensembles | | | | | | |
|---|---|---|---|---|---|---|---|
| | **Ours** | DES-P | KL pruning | DFP | MDM pruning | Meta pruning | META-DES.O |
| Adult | 10.22 | 11.36 | 14.76 | 18.56 | 8.67 | 11.16 | 19.30 |
| Arcene | 11.73 | 18.63 | 20.83 | 28.56 | 19.10 | 21.21 | 18.76 |
| Australian | 11.40 | 15.83 | 14.76 | 24.43 | 8.50 | 11.73 | 12.83 |
| Breast-cancer | 8.43 | 9.46 | 26.10 | 21.03 | 7.80 | 9.82 | 10.20 |
| Disorders | 12.70 | 24.70 | 24.58 | 22.16 | 17.71 | 13.96 | 24.10 |
| Heart | 9.43 | 32.43 | 17.94 | 15.06 | 13.60 | 11.93 | 14.16 |
| Wine | 2.98 | 13.90 | 5.75 | 3.90 | 5.98 | 4.11 | 3.70 |
| Ionosphere | 5.45 | 13.53 | 10.50 | 6.47 | 10.73 | 8.46 | 6.47 |
| Kr-vs-kp | 4.29 | 35.76 | 10.61 | 8.93 | 7.20 | 10.12 | 9.40 |
| Letter | 6.35 | 23.59 | 7.33 | 5.24 | 11.23 | 10.53 | 7.46 |
| Optdigits | 23.57 | 22.46 | 25.16 | 23.13 | 23.76 | 25.80 | 23.06 |
| Satimage | 17.73 | 38.03 | 22.26 | 24.20 | 25.96 | 22.91 | 22.90 |
| Sonar | 11.34 | 13.68 | 31.06 | 16.73 | 34.95 | 20.70 | 23.16 |
| Spambase | 17.54 | 23.90 | 23.33 | 26.93 | 36.60 | 21.38 | 26.93 |
| Tic-tac-toe | 18.32 | 11.86 | 21.26 | 11.10 | 33.23 | 15.63 | 9.76 |
| Vehicle | 16.54 | 15.96 | 17.73 | 24.00 | 27.36 | 24.42 | 31.53 |
| Vote | 2.46 | 3.57 | 5.14 | 3.17 | 5.45 | 6.02 | 2.90 |
| **Count of wins** | 10 | 2 | 0 | 1 | 3 | 0 | 1 |
| **Direct wins** | | 14 | 17 | 14 | 14 | 16 | 15 |

## 4. Real-world problems

### 4.1 Uncertainty quality metrics assessment by benchmarking

TABLE 4: Three different uncertainty prediction metrics applied to nine regression benchmarking datasets

| | Loss | | | PICP | | | MPIW | | |
|---|---|---|---|---|---|---|---|---|---|
| | **Ours** | CD | LUBE | **Ours** | CD | LUBE | **Ours** | CD | LUBE |
| Boston | 1.04±0.09 | 1.76±0.28 | 1.33±0.05 | 0.81±0.01 | 0.89±0.02 | 0.92±0.01 | 0.48±0.03 | 0.87±0.03 | 1.16±0.02 |
| Concrete | 1.06±0.02 | 1.23±0.06 | 1.16±0.02 | 0.90±0.00 | 0.92±0.01 | 0.94±0.01 | 0.71±0.04 | 1.00±0.02 | 1.09±0.01 |
| Energy | 0.34±0.01 | 0.50±0.02 | 0.47±0.01 | 0.95±0.01 | 0.99±0.00 | 0.97±0.01 | 0.49±0.01 | 0.50±0.02 | 0.47±0.01 |
| Kin8nm | 0.58±0.02 | 1.14±0.01 | 1.24±0.01 | 0.97±0.00 | 0.97±0.00 | 0.96±0.00 | 1.12±0.01 | 1.14±0.01 | 1.25±0.01 |
| Naval | 0.18±0.02 | 0.31±0.01 | 0.27±0.01 | 1.00±0.00 | 0.99±0.00 | 0.98±0.00 | 0.48±0.01 | 0.31±0.01 | 0.28±0.01 |
| Power | 0.30±0.01 | 0.91±0.00 | 0.86±0.00 | 0.96±0.00 | 0.96±0.00 | 0.95±0.00 | 0.88±0.02 | 0.91±0.00 | 0.86±0.00 |
| Wine | 3.25±0.21 | 4.13±0.31 | 3.13±0.19 | 0.84±0.01 | 0.90±0.01 | 0.92±0.01 | 1.21±0.02 | 2.50±0.02 | 2.33±0.02 |
| Yacht | 0.10±0.01 | 0.31±0.02 | 0.23±0.02 | 0.88±0.01 | 0.98±0.01 | 0.96±0.01 | 0.09±0.01 | 0.30±0.02 | 0.17±0.00 |
| Year | 2.04±NA | 2.90±NA | 2.47±NA | 0.98±NA | 0.96±NA | 0.96±NA | 2.60±NA | 2.91±NA | 2.48±NA |

To measure the quality of uncertainty prediction, MBPEP is benchmarked on nine UCI regression datasets. Due to different loss functions and model architectures having been used in previous models, it is unfair to directly compare the performance among these models. As a PI model, MBPEP uses a unique loss function to shrink the width between the upper and lower bounds. Other loss functions include the Concrete Dropout model (CD) [34] that has been described in Section 3.2.1. To convert the mean and variance into the metrics PICP and MPIW, according to [23], the mean and standard deviation of the uncertainty are set equal to $(\hat{Y}_U - \hat{Y}_L)/2$ and $(\hat{Y}_U - \hat{Y}_L)/3.98$ respectively. As for LUBE model [15] that also has been described in Section 3.2.1, in contrast to our MBPEP, it only uses fixed activation functions (the Sigmoid function in the paper), and soft activations are not involved in constructing the MPIW.

Three quality assessment metrics learning losses, PICP and MPIW are employed. The results are shown in Tab. 4. Each result is recomputed 20 times and is presented in the form of mean ± standard error . The number of hidden units of each base learner is set to 100 , and the inputs are normalized to [0,1] . The loss metric in Tab. 4 shows that our model achieves the minimum learning losses on most of datasets. Additionally, our model also achieves the narrowest interval width (the minimum value of *MPIW* ) on most of these datasets. As to the coverage probability metric (PICP), our loss function is comparable to the other two losses.

### 4.2 Real-world classification on MNIST and SVHN

In this section, MBPEP is applied to evaluate its uncertainty classification performance on the real-world datasets. The training dataset is MNIST [39], and the test dataset is SVHN [40]. MNIST is a standard digital character benchmark dataset that contains 60000 and 10000 of digits, while SVHN is composed of letters of the alphabets instead of digits. As these two datasets have different data distributions, this experiment can uncover our method's uncertainty classification ability. In particular, each image in SVHN needs to be resized to the same size as that of MNIST images. Fig. 5 shows several images from these two datasets.

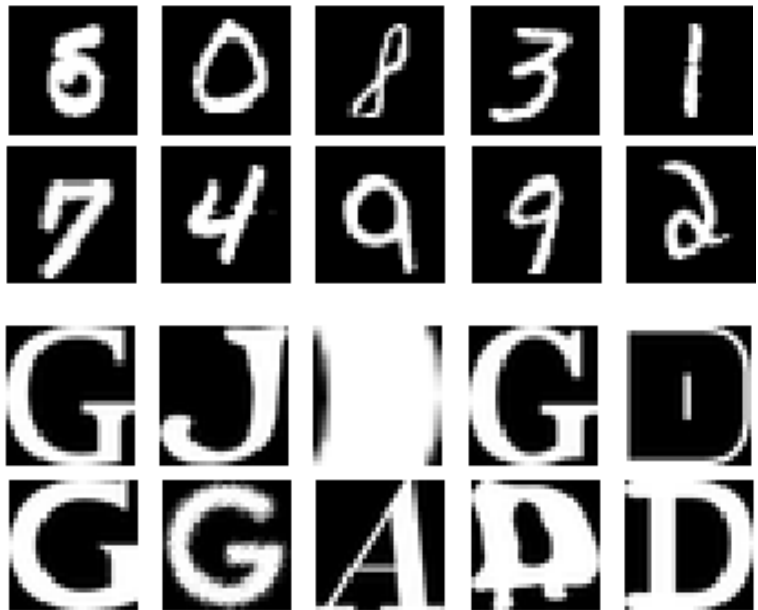

**Fig. 5**: (a). Examples of handwritten digits from MNIST (b). Examples of images of letters from SVHN

The classical Lenet-5 architecture, which consists of 2 convolutional layers, 2 pooling layers and a fully connection layer, is used as the base leaner in this experiment. MBPEP is compared to state-of-the-art models such as Deep ensembles with adversarial training (DE) [23] and MC-dropout [10] on 1000 samples of MNIST and SVHN. Due to the ground truth labels belonging to $[0,9]$, when the known dataset (MNIST) inputs are used in learning, outputs will show large values of probabilities, and when the unknown dataset (SVHN) inputs are processed, the confidence degrees of probabilities will shrunk. We calculate the conditional entropies of output predictive probabilities for unique labels to evaluate the quality of uncertainty estimation.

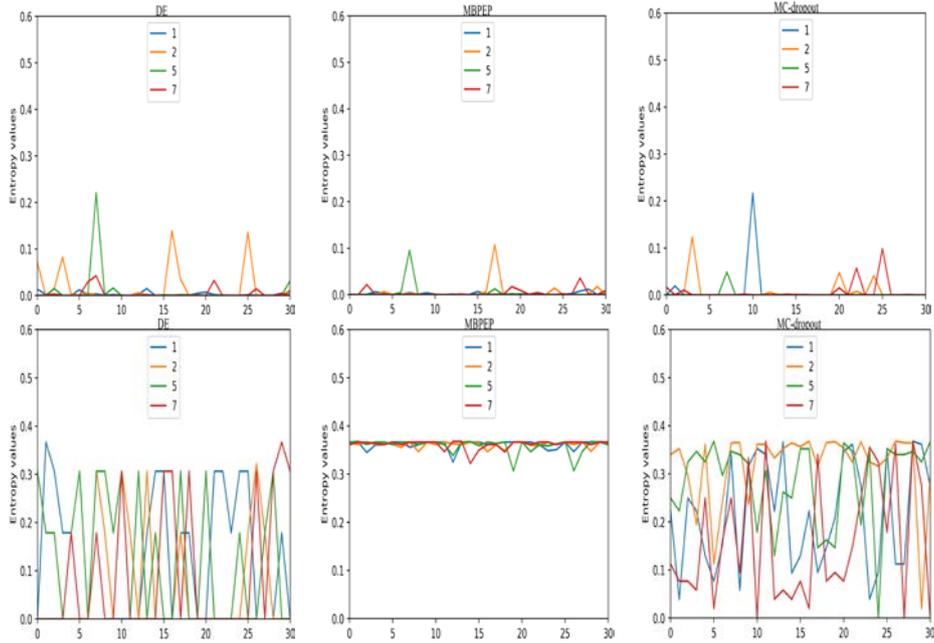

**Fig. 6**: Distributions of the entropy values for three different models. The panels from left to right show the performances of DE, MBPEP and MC-dropout. The top row of these panels shows the test examples from the known dataset (MNIST), and the bottom row shows the test examples from the unknown dataset (SVHN).

Due to 1000 samples of MNIST and SVHN being too many for one figure, the entropy values of 30 samples for labels 1, 2, 5 and 7 are shown in Fig. 6. The top row shows that all three models achieve the lowest entropy values, and the performance of MBPEP is more stable than that of other methods. As to the uncertainty prediction performance in the bottom row of Fig. 6, the results show that MBPEP achieves the highest uncertainty degrees. Although DE and MC-dropout also exhibit some uncertainty degrees, their output entropy values show high variances. These results indicate that MBPEP is a superior model for handling uncertainty tasks.

### 4.3 Multitask problem

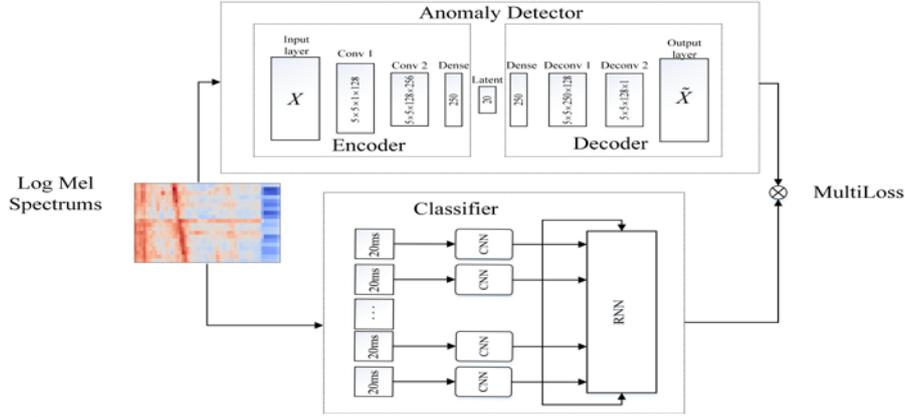

**Fig. 7**: Applying uncertainty estimation to a multitask learning scene. The log mel spectra are feed as inputs. The joint learning of the convolutional autoencoder and the stack convolution recurrent network deals with the novelty detection and acoustic classification tasks.

In this section, MBPEP is applied into the multitask uncertainty learning that aims to enhance the performance with respect to the IOT acoustic task. The training dataset contains 5140 positive samples belonging to 3 labels; the testing dataset contains 2999 negative samples that can be considered as 'novelty', and 509 positive samples belonging to 3 labels that are the same as those in training dataset. We aim to simultaneously detect the novelty samples and classify the remaining positive samples. The jointly learning method of novelty detection and classification tasks is proposed. Fig. 7 shows our multitask learning framework that incorporates the log mel spectra [41] and has two pipelines. In the training phase, the convolutional autoencoder only rebuilds the space of the positive samples while the novelty samples can not been effectively reconstructed, and Stacked CRNN [42] classifier is responsible for learning the determination boundaries of every positive samples. In the testing phase, the convolutional autoencoder is used for novelty detector and Stacked CRNN predicts the positive samples that are filtered by the autoencoder. As to the multi-task loss function, the multiple outputs of MBPEP models are with the $CVAE$ output and $CRNN$. Hence, we construct the loss function of the jointly uncertainty learning similarly to [43]:

$$Loss_{multi} = W_{CVAE} * Loss_{CVAE} + W_{CRNN} * Loss_{CRNN} \qquad (12)$$

where $Loss_{CRNN}$ is the Softmax uncertainty likelihood function [40], and $Loss_{CVAE}$ is the uncertainty loss function in Eq. (7). In Table 5, several experiments are presented to show the performance of our multitask learning. The accuracy and the precision/recall are used as model metrics to evaluate the performance of detection and classification.

To quantitatively analyze the performance of multitask learning, the optimal weights which are predicted by the MBPEP to predefined weights for each independent task. In Tab. 5, the performance of each independent model and multitask learning model with optimal weights or predefined weights is shown. (Note: "De."

denotes detection, and "Cla." denotes classification weights in Tab. 5) It is observed that if combined loss of two tasks is modeled by jointly uncertainty MBPEP learning, the corresponding detection accuracy and precision/recall of classification can be improved.

TABLE 5: Quantitative comparison of performance of uncertainty predicted weights to predefined learning novelty detection and classification

| Loss | Task weights De. | Task weights Cla. | Detection Accuracy(%) | Classification Precision/Recall |
|---|---|---|---|---|
| Detection only | 1 | 0 | 51.88% | 0.80, 0.82 |
| Classification only | 0 | 1 | 53.58% | 0.82, 0.85 |
| Average weights | 0.5 | 0.5 | 53.53% | 0.87, 0.89 |
| **Uncertainty predicted Optimal weights** | 0.34 | 0.66 | 60.05% | 0.93, 0.93 |
| 2 task uncertainty weighting | 0.34 | 0 | 54.77% | 0.81, 0.86 |
| 2 task uncertainty weighting | 0 | 0.66 | 57.21% | 0.85, 0.91 |

## 5. Conclusions

In this manuscript, the margin-based Pareto deep ensemble pruning (MBPEP) model is derived to deal with the uncertainty estimation problems. Consisting of several sub-learners pruned by the modified Pareto algorithm, MBPEP achieves a small mean prediction interval width (MPIW) with a large prediction interval coverage probability (PICP). Specifically, the loss function of each sub-learner is reconstructed to link MPIW and PICP, and more nonlinear operators are added to allow the sub-leaner to be easily solved by the standard gradient descents.

Various loss functions and ensemble pruning methods are compared; the results of the experiments show that MBPEP can achieve a narrow interval width, reducing the testing time consuming and optimal ensembles under low learning errors, which demonstrates that MBPEP is a powerful tool for uncertainty estimation. It can output a high uncertainty degree when facing the real-world unknown data distribution problem (MNIST and SVHN).

Understandably, MBPEP also has a weakness. In Tab. 4, its metric PICP is not clearly superior to others methods'. In the future research, we will discuss the relationship between PICP and MPIW and continue to explore advancing the model to deal with the uncertainty estimation problems.

## Acknowledgments

This work was supported by the National Natural Science Foundation of China (61874079, 61574102 and 61774113), the Fundamental Research Fund for the Central Universities, Wuhan University (2042018gf0045, 2042017gf0052), the Wuhan Research Program of Application Foundation (2018010401011289), and the Luojia Young Scholars Program. Part of calculation in this paper has been done on the supercomputing system in the Supercomputing Center of Wuhan University.